\documentclass[11pt,a4paper]{article}

\usepackage[utf8]{inputenc}
\usepackage[english]{babel}

\usepackage[square,sort,comma,numbers]{natbib}
\usepackage{graphicx}
\DeclareGraphicsExtensions{.pdf,.png,.jpg}

\usepackage{algorithm}
\usepackage{algpseudocode}

\usepackage{tikz}
\usetikzlibrary{calc,intersections,through,backgrounds}

\usepackage{pgfplots}

\usepackage{authblk}
\newcommand{\specialcell}[2][c]{%
    \begin{tabular}[#1]{@{}c@{}}#2\end{tabular}}

\title{Large scale near-duplicate image retrieval using Triples of Adjacent Ranked Features (TARF) with embedded geometric information}
\author[1]{Sergei Fedorov\thanks{\texttt{sfedorov@abbyy.com}}}
\author[1]{Olga Kacher\thanks{\texttt{okacher@abbyy.com}}}
\affil[1]{ABBYY, Moscow, Russia}
\date{\today}

\begin{document}

\maketitle
\begin{abstract}
Most approaches to large-scale image retrieval are based on the construction of the inverted index of local image descriptors or visual words. A search in such an index usually results in a large number of candidates. This list of candidates is then re-ranked with the help of a geometric verification, using a RANSAC algorithm, for example. In this paper we propose a feature representation, which is built as a combination of three local descriptors. It allows one to significantly decrease the number of false matches and to shorten the list of candidates after the initial search in the inverted index. This combination of local descriptors is both reproducible and highly discriminative, and thus can be efficiently used for large-scale near-duplicate image retrieval.
\end{abstract}

\section{Introduction}

Most systems for large-scale image retrieval are designed for a scenario in which a query image is searched for within a large image set (hundreds of thousands or millions of images), and the goal is to find its partial duplicates. To solve this task most state-of-the-art algorithms make use of the Bag of Words (BoW) representation of an image, where each local descriptor extracted from an image is represented with a visual word, and implement an effective search in the inverted index (see for example~\cite{Sivic2003video-google}). One of the problems of all such algorithms is the large list of initial candidates, which needs to be further pruned with geometrical checks, usually with a RANSAC algorithm. For large image sets the candidate list can be very long, which makes it infeasible to perform geometric verification for each candidate. In order to increase the discriminative power of the features extracted from an image, and to decrease the size of the initial candidates list, different ways of embedding spatial information into the features were suggested. Among those are bundling features~\cite{Wu2009bundling-features}, bundle min-hashing~\cite{Romberg2013BundleMinHashing}, nested SIFT~\cite{Xu2013nested-sift}, geometry-preserving visual phrases~\cite{Zhang2011visual-phrases}, and weak geometrical consistency~\cite{Jegou2008hammingEmbedding}.
These approaches allow one to significantly increase the effectiveness of the duplicates search in a large image set.

For the closely related task of logo recognition there have also been proposed methods for adding geometric information to the inverted index. In~\cite{Romberg2014Aggregating} triples of SIFT features were used for logo detection, and in~\cite{Kalantidis2011ScalableTriangulationBased} authors proposed multi-scale Delaunay triangulation for grouping local features into triples. In both papers~\cite{Romberg2014Aggregating,Kalantidis2011ScalableTriangulationBased} a set of training images with the same logo is required for the construction of a logo model, a key point by which these approaches differ from ours, in which a single image is used for extracting triples of local features and indexing.

In this paper we consider the task of finding duplicates between two large image sets, both with hundreds of thousands or millions of images. One possible approach is to build a search index using the first image set, and then perform the search within it for each image from the second image set using one of the methods discussed above. This approach requires a large number of search queries, and is absolutely infeasible when the search index does not fit in the main memory. The problem here is the large list of initial candidates. For example, nested SIFT would first generate the list of all images with matched bounding features, and only then this list will be pruned by the requirement that there should be similar member features. Finding duplicates between two large image sets is in fact equivalent to considering all possible pairs of images, thus the initial list of candidates would be vast.

To address this problem we propose a composite local feature, which we have named TARF. A TARF descriptor is a combination of three local descriptors, and because of this it is highly discriminative, thus it allows us to significantly decrease the number of false matches and to shorten the list of candidates after the initial search in an inverted index. In order to increase the reproducibility of TARFs we propose a method of their extraction which takes into account the scores of constituent feature points, described in detail in the next section. Adding geometric information to this feature is straightforward and further increases its discriminative power, allowing us to achieve a very low false positives rate. This makes it possible to use TARF descriptors for the task of finding duplicates between two image sets even for very large image sets that don't fit in the main memory.

\section{TARF features}
In this paper we propose a new composite local feature, Triple of Adjacent Ranked Features (TARF). These triples of feature points are used to construct complex descriptors, which include three local descriptors,
as well as the geometric layout of the three points. It is essential that these complex descriptors should be a) highly distinctive, and b) reproducible. To achieve these aims, we propose the following
scheme for extracting TARFs from an image.

\subsection{Extracting TARFs}
Triples contain heterogeneous feature points. One feature point is detected with a blob detector, such as SIFT~\cite{Lowe2004sift} or SURF~\cite{Bay2008surf}. Two other points are found with a corner detector, such as the BRISK detector~\cite{Leutenegger2011brisk}. We will denote these points as BLOB and CORNER. A group of three points gives a richer description of the local image area, as compared to a single feature point, and includes very distinctive geometric features.
We first extract BLOB feature points from an image, and then each BLOB feature point is used to select nearby CORNER feature points. Enumerating all such triples would lead to a large number of combinations, most of which would not be very reproducible, because even if one of the three points is missing on a duplicate image, the whole triple will be lost. So, it is essential to select only those triples that have a good chance of being reproduced on a duplicate. To achieve this, we make use of the score of CORNER points (given by the CORNER points detector), modified to take into account position of the CORNER points relative to a BLOB feature point.

More precisely, the modified score is calculated as
\begin{equation}
\left\{
\begin{array}{l}
S^* = \exp\left[ -0.5 \left( \frac{d - d_0}{\sigma_d} \right)^2 \right] \exp\left[ -0.5 \left( \frac{R_c - R_0}{\sigma_R} \right)^2 \right] S \\
d_0 = 0.5 R_b, \quad \sigma_d = 0.15 R_b \\
R_0 = 0.33 R_b, \quad \sigma_R = 0.15 R_b
\end{array}
\right.
\label{eq:BriskModifiedScore},
\end{equation}
where $R_c$ and $R_b$ are CORNER and BLOB feature points radii correspondingly, $d$ -- distance between centers of CORNER and BLOB, $S$ and $S^*$ are original and modified CORNER scores.
Parameters $d_0$, $\sigma_d$, $R_0$ and $\sigma_R$ were determined experimentally; the values in~(\ref{eq:BriskModifiedScore}) correspond to TARFs based on SIFT/BRISK.
These parameters play an important role for the quality of the TARF extractor.
All CORNER points which lie in the vicinity of a BLOB feature point are sorted by modified score $S^*$, and $N_{top}$ CORNER feature points are taken.
The value $N_{top} = 7$ was found to give best results.
Next a global score threshold $S_0^*$ is determined in such a way that taking all CORNER points with modified score $S^* < S_0^*$ gives a total of $N_0$ triples:

\begin{equation}
N_0 = \sum\limits_{i \in \mbox{all BLOB points}} n_i (n_i - 1) / 2
\label{eq:BriskModifiedScoreThreshold},
\end{equation}

where $n_i$ is the number of CORNER points in the vicinity of $i$-th BLOB, with modified score below threshold, $S^* < S_0^*$.
$N_0$ is a parameter of the detector, usually $N_0 = 2500 .. 3000$

The exact algorithm for extracting TARF features:
\begin{algorithm}[H]
\caption{Extracting TARFs}
\begin{algorithmic}[1]
\State Extract BLOB feature points with BLOB detector
\State Extract CORNER feature points with low threshold (for BRISK detector threshold 5 was used)
\For{each BLOB point}
\State Choose all CORNER points from the neighbourhood of BLOB feature point
\State Re-rank CORNER points according to~(\ref{eq:BriskModifiedScore})
\State Choose top $n_*$ CORNER points ($n_*=7$)
\EndFor
\State Find global threshold for the scores of CORNER points. For each BLOB select $n_i$ top CORNER feature points in its neighbourhood with the score above global threshold, $n_i \leq n_*$ \label{alg_find_global_threshold}
\State Add all combinations (BLOB point + 2 different CORNER points from the list, determined in step~\ref{alg_find_global_threshold}) to the list of extracted TARFs
\end{algorithmic}
\end{algorithm}

\subsection{Matching TARFs}

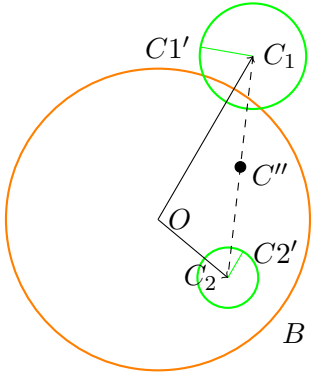
\begin{figure}
\begin{tikzpicture}
\coordinate[label=right:$O$] (O) at (4, 2);
\draw[orange,thick] (O) circle(2cm);
\coordinate[label=right:$B$] (label1) at (5.5, 0.5);

\coordinate[] (C1) at ($ (O) + (60:2.5) $);
\draw[green,thick] (C1) circle(0.7cm);
\draw[green,thin] (C1) -- ++(170 : 0.7cm) node[anchor=east, black]{$C1'$};
\draw[->,black,thin] (O) -- ++(60:2.5);
\coordinate[label=right:{$C_1$}] (label3) at (C1);

\coordinate[] (C2) at ($ (O) + (320:1.2) $);
\draw[green,thick](C2) circle(0.4cm);
\draw[green,thin] (C2) -- ++(60 : 0.4cm) node[anchor=west, black]{$C2'$};
\draw[->,black,thin] (O) -- (C2);
\coordinate[label=left:{$C_2$}] (label2) at (C2);

\coordinate[] (Cprime2) at ($ (C1) !.5! (C2) $ );
\filldraw[] (Cprime2) circle (2pt);
\coordinate[label=right:{$C''$}] (label5) at (5.1, 2.6);

\draw[dashed,black,thin] (C1) -- (C2);

\end{tikzpicture}

\caption{TARF descriptor. The descriptor is composed of BLOB point $B$ with the center at $O$ and radius $R_b$ and two CORNER points, $C_1$ and $C_2$.
Segments $C_1 C_1'$ and $C_2 C_2'$ are related to the directions of CORNER feature points.
$C''$ is the middle point of the line segment $C_1 C_2$.
}
\label{fig_TARF_scheme}

\end{figure}

\begin{figure}
\includegraphics[height=5cm]{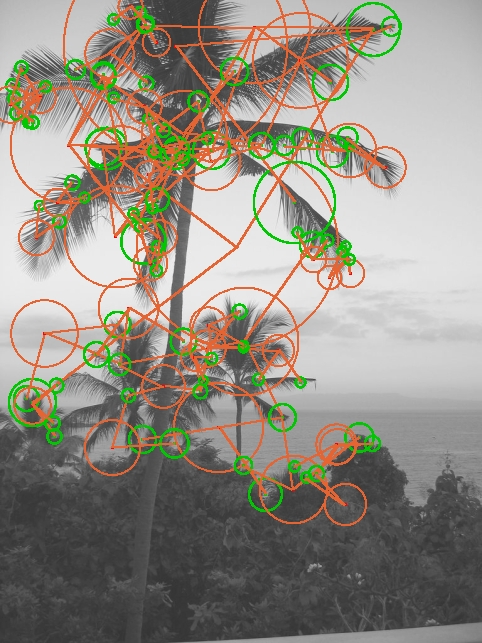}
\hspace{0.5cm}
\includegraphics[height=5cm]{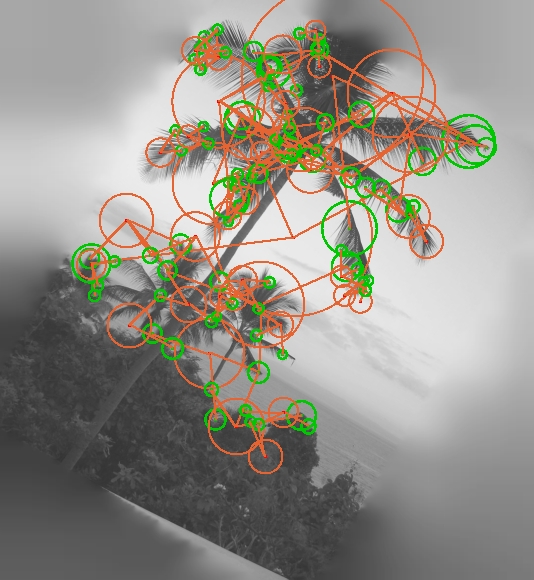}
\caption{Matched TARF features on a partial duplicate images with 30 degrees rotation}
\label{fig_TARF_example}
\end{figure}

Matching TARF features is straightforward: descriptors of all three constituent points should be matched, and the geometrical layout of three feature points should be similar.
Geometrical features describing the TARF layout may be chosen in different ways. Our choice of geometrical features is given below ($R_b$ is the radius of the BLOB feature point):
\begin{enumerate}
\item Angle between vectors $OC_1$ and $OC_2$, $\alpha = \angle( C_1 O C_2)$
\item Angle between vectors $OC_1$ and $C_1 C_1'$, $\beta_1 = \angle( O C_1 C_1')$
\item Angle between vectors $OC_2$ and $C_2 C_2'$, $\beta_2 = \angle( O C_2 C_2')$
\item Ratio $\epsilon_1 = O C_1 / O C_2$
\item Ratio $\epsilon_2 = C_1 C_2 / R_b$
\item Ratio $\epsilon_3 = O C'' / R_b$
\end{enumerate}
Fig.~\ref{fig_TARF_scheme} shows the geometric layout of feature points in TARF. An example of partial duplicates with matched TARF descriptors is given in Fig.~\ref{fig_TARF_example}.

\subsection{Matching TARFs in the inverted index}
Two TARFs are fully matched if all the three constituent descriptors are matched and their geometric layout is similar. Geometric layout is checked by verifying that all the geometric features listed above differ by less than a threshold:
\begin{equation}
\left\{
\begin{array}{l}
| \alpha^{(1)} - \alpha^{(2)} | < t_1, \\
| \beta_k^{(1)} - \beta_k^{(2)} | < t_1, \quad k = 1, 2\\
| \epsilon_k^{(1)} - \epsilon_k^{(2)} | < t_{\epsilon}, \quad k = 1, 2, 3\\
\end{array}
\right.
\label{eq:geometricLayoutChecks}
\end{equation}
Thresholds were determined experimentally, $t_1 = 20^{\circ}$, $t_{\epsilon} = 0.1$.

\subsubsection{Matching descriptors with visual words}
For large scale image retrieval it's essential to be able to match three constituent descriptors of TARFs simultaneously. The descriptors are represented in the inverted index with visual words. Three visual words, assigned to three descriptors, are packed into one 32-bit integer. Constituent descriptors of TARFs are matched if these 32-bit integers are exactly equal. Moreover, each visual word has an inverse document frequency (IDF) score associated with it. Several visual words with the lowest score (i.e. belonging to the largest clusters of descriptors) are placed into the stop list, and TARFs with such constituent descriptors are ignored.

One of the problems is that such a representation with visual words makes the probability of matching two similar descriptors significantly lower, as compared to matching with e.g. LSH (locality sensitive hashing), even for small vocabulary sizes. To illustrate this, Table~\ref{table_reproducibility} shows experimental results for probabilities of matching descriptors by various methods. For a set of 500 images we generated transformed duplicates; transformations included rotations and downscale. Feature points were extracted from original images and from duplicates.
Those feature points that are at the same position and scale (up to image transformation) are considered to be reproduced. The reproducibility rate $R$ is the percentage of feature points that were extracted from the original image and were reproduced on the transformed image. Depending on the image transformations this rate is $R = 30..40\%$ for both SIFT and BRISK detectors.
Next, pairs of descriptors were collected for the reproduced feature points, taking one descriptor from the original image, and another from the transformed image. These pairs of descriptors were matched using various methods, and the percentage of matched pairs is given in the TPR column. Thus, the overall probability of finding a match for a particular feature point from the original image on a transformed image is given by $p = R\cdot \mbox{TPR}$.
The probability of false matches is given in the FPR column, i.e FPR is the match rate for pairs of descriptors calculated at random uncorrelated points of images.

As seen from Table~\ref{table_reproducibility}, matching with visual words results in a TPR that is lower compared to matching by threshold, even for small vocabulary sizes. Though usually this TPR is sufficient, sometimes it may be desirable to increase the recall. This can be achieved by training two different vocabularies for each type of descriptor, and representing each descriptor with two visual words (different vocabularies are trained using clusterization of descriptors extracted from different sets of images). In this case descriptors are matched if any of the two visual words is matched. This makes TPR significantly higher and comparable with matching by a threshold, though FPR is of course much higher (see Table~\ref{table_reproducibility}).

If this approach with several vocabularies of visual words is taken, then each TARF descriptor will be represented by $2\times 2\times 2 = 8$ 32-bit integers. This requires 8 tables in the inverted index. Descriptors from two TARFs are considered to be matched if any of the eight 32-bit integers matches exactly. This can be viewed as a modification of LSH, with different hashes being generated by different vocabularies of visual words. Other TARF representations may also be considered: a) a BLOB descriptor is represented with 1 visual word, and a CORNER descriptor is represented with 2 visual words ($1\times 2\times$ 2) or b) 2 visual words for BLOB, and 1 visual word for CORNER, $2\times 1\times 1$.

Equation~(\ref{eq:SeveralVisualWords}) illustrates how a TARF descriptor can be represented with visual words when using two vocabularies for each type of constituent descriptor.

\begin{equation}
\begin{array}{l}
\left\{
\begin{array}{l}
     \mbox{BLOB point descriptor}\to (w^b_1, w^b_2) \\
     \mbox{1st CORNER point descriptor}\to (w^{c_1}_1, w^{c_1}_2) \\
     \mbox{2nd CORNER point descriptor}\to (w^{c_2}_1, w^{c_2}_2)
\end{array}
\right.
\\  \\
\begin{array}{l}
\mbox{TARF}\to (w^b_1, w^b_2) \times (w^{c_1}_1, w^{c_1}_2) \times (w^{c_2}_1, w^{c_2}_2) = \\
   ( \{ w^b_1, w^{c_1}_1, w^{c_2}_1 \}, \{ w^b_1, w^{c_1}_1, w^{c_2}_2 \}, ..., \{w^b_2, w^{c_1}_2, w^{c_2}_2 \} )
\end{array}
\end{array}
\label{eq:SeveralVisualWords}
\end{equation}

\begin{table}
\begin{tabular}{c}
\begin{tabular}{|c|c|c|}
\hline
matching method & FPR & TPR \\
\hline
threshold, L2 distance 0.4 & $3\cdot10^{-4}$ & 0.75 \\
visual words, 128 words & $8.5\cdot10^{-3}$ & 0.65 \\
visual words, 256 words & $4.3\cdot10^{-3}$ & 0.61 \\
visual words, 512 words & $2.2\cdot 10^{-3}$ & 0.57 \\
visual words, 2 groups of 128 words & $1.3\cdot10^{-2}$ & 0.76 \\
visual words, 2 groups of 256 words & $6.6\cdot10^{-3}$ & 0.73 \\
visual words, 2 groups of 512 words & $3.4\cdot 10^{-3}$ & 0.7 \\
\hline
\end{tabular} \\
SIFT \\ \\

\begin{tabular}{|c|c|c|}
\hline
matching method & FPR & TPR \\
\hline
threshold, hamming distance 90 & $7\cdot10^{-4}$ & 0.57 \\
visual words, 128 words & $10^{-2}$ & 0.43  \\
visual words, 256 words & $5.2 \cdot 10^{-3}$ & 0.38 \\
visual words, 512 words & $2.8 \cdot 10^{-3}$ & 0.35 \\
visual words, 2 groups of 128 words & $1.9\cdot 10^{-2}$ & 0.6 \\
visual words, 2 groups of 256 words & $9.6\cdot 10^{-3}$ & 0.54 \\
visual words, 2 groups of 512 words & $5.1\cdot 10^{-3}$ & 0.5 \\
\hline
\end{tabular} \\
BRISK\\ \\

\end{tabular}
\caption{Descriptor match rates.}
\label{table_reproducibility}
\end{table}

\subsection{TARFs reproducibility and match rates}
We have tested TARF features on the same set of 500 images with transformed duplicates as described in the previous section. TARFs that were extracted in the same places, i.e. in which all three constituent feature points were found at the same positions and scale (up to transformations), were considered to be reproduced. The reproducibility of TARFs is $R_{T} = 8\%$. Though this is a low number, it is still much higher than the reproducibility of randomly chosen triples of feature points, and this is due to the proposed scheme of TARFs extraction.

\begin{table}
\begin{tabular}{|c|c|c|}
\hline
matching method & FPR & TPR \\
\hline
threshold, geometric features discarded & $2.5\cdot 10^{-5}$ & 0.7 \\
\hline
threshold & $4\cdot 10^{-8}$ & 0.68 \\
visual words, 1x1x1, no stop list & $4 \cdot 10^{-9}$  & 0.28 \\
visual words, 1x1x1, stop lists of size 10 & $1.5 \cdot 10^{-9}$ & 0.25 \\
visual words, 2x1x1, stop lists of size 10 & $2.5 \cdot 10^{-9}$ & 0.27 \\
visual words, 1x2x2, stop lists of size 10 & $4 \cdot 10^{-9}$ & 0.41 \\
visual words, 2x2x2, stop lists of size 10 & $5 \cdot 10^{-9}$ & 0.44 \\
\hline
\end{tabular}
\caption{TARF match rates. First row corresponds to matching without checks of geometric layout. Other rows correspond to fully matched TARFs.}
\label{table_reproducibility_tarf}
\end{table}

Table~\ref{table_reproducibility_tarf} shows the FPR and TPR for TARFs. The probability of false matches is given in the FPR column. TPR is defined as the probability that a pair of reproduced TARF descriptors is matched.

Normally TARFs are matched fully, including checks of geometric layout~(\ref{eq:geometricLayoutChecks}). In order to evaluate the importance of geometric features, we have tested TARFs matching without performing these checks, i.e. discarding geometric features and matching only constituent decriptors. Results are presented in the first row of the table. It can be seen that discarding geometric features leads to an FPR that is several orders of magnitude higher than the FPR for full TARF matching, yet it is still quite low at $2.5\cdot 10^{-5}$. Including geometric features makes TARFs highly discriminative, with the FPR as low as $10^{-9}..10^{-8}$. Using stop lists for the visual words further reduces FPR, making it several times lower, and almost does not affect TPR.
The TPR depends on the matching method; it has the largest value of $0.68$ if the constituent descriptors are matched by threshold. In the case of matching constituent descriptors with visual words it is lower and varies from $0.25$ to $0.44$; the TPR is higher when several dictionaries of visual words are used.

\section{Experimental results}

The search of near-duplicate images is performed as follows. First, a search index is built. It is an inverted index of three visual words packed into 32-bit integers. Geometric data is stored in the index as auxiliary data. Three IDF scores of constituent visual words are summed up to give the IDF score of the TARF.

After finding all the images that have at least one TARF matching a query image, the score of the candidate images is calculated as the sum of IDF scores of all matched TARFs. A precision-recall curve can be obtained by varying the threshold for this score. Matched images can be filtered by applying a geometric model. We use a standard RANSAC algorithm. RANSAC converges very fast, because of the extremely low number of outliers, so we make only 10 iterations of RANSAC.

An important property of TARFs is a very low false positive rate. Table~\ref{table_tarf_fp_probability} shows the probability of finding a given number of matched TARF features for a pair of non-duplicate images. It can be seen that the probability decreases fast as the the number of false positive matches increases. It should be noted, that the use of stop lists for visual words is essential and reduces the number of false candidate images by the order of magnitude. If one builds a search index with 100K images using the 1x1x1 scheme with stop lists of size 10, and then searches for a query image in this index, the result of such a search will have on average as little as 400 false candidate images with one matched TARF descriptor, 10 false candidates with two matched TARFs and 1 false candidate with three matched TARFs.
So, checking all the candidates with more sophisticated algorithms, e.g. with geometric verification by a RANSAC algorithm, is feasible even for very large search indices.

\begin{table}
\begin{tabular}{|c|c|c|}
\hline
number of matched TARFs & \multicolumn{2}{|c|}{ probability } \\
\cline{2-3}
 & \specialcell[c]{ 1x1x1 scheme\\ no stop lists } & \specialcell[c]{ 1x1x1 scheme\\ stop lists of size 10 } \\
\hline
0 & 0.985 & 0.995 \\
\hline
1 & 0.013 & 0.004 \\
2 & $10^{-3}$ & $10^{-4}$ \\
3 & $3\cdot10^{-4}$ & $10^{-5}$ \\
\hline
\end{tabular}
\caption{Probability of a given number of matched TARFs on a pair of non-duplicate images}
\label{table_tarf_fp_probability}
\end{table}

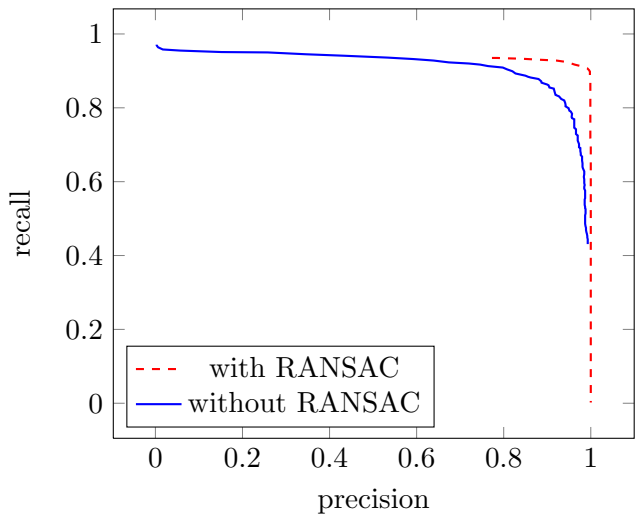
\begin{figure}
\begin{tikzpicture}
\begin{axis}[
	xlabel={precision},
	ylabel={recall},
    legend style={at={(0.025, 0.025)}, anchor=south west},
]
\addplot[no markers, red, dashed, thick] table[x=precision, y=recall] {precision-recall-30k-ransac.txt};
	
\addplot[no markers, blue, thick] table[x=precision, y=recall] {precision-recall-30k-noransac.txt};
\legend{
    with RANSAC,
    without RANSAC
}
\end{axis}
\end{tikzpicture}
\caption{
The effect of geometric verification with RANSAC on the precision-recall curve. Query images are duplicates downscaled to 30k pixels; original images are diluted with 100K distracting images.
}
\label{fig_precision_recall}
\end{figure}

In order to evaluate image retrieval based on TARFs we have constructed an image set of 1853 images, and four sets of near-duplicates of these images, with the following modifications: a) downscale to small size ~30k pixels (150x200), b) rotation 30 degrees, c) crop 70\% (30 \% of an image retained) and d) strong gaussian blur with $\sigma=4$. Original images were added to the search index and diluted with the 100K distracting images, taken from
the Flickr 100k Dataset~\cite{Philbin2007}. For each of these modifications we have evaluated the precision-recall curve, and have measured AP (average precision), defined as the area under the precision-recall curve. Figure~\ref{fig_precision_recall} illustrates how geometric verification with RANSAC affects the precision-recall curve.

We have varied different parameters of the image retrieval system, and measured the resulting performance. The default
parameters were: BLOB -- SIFT descriptors, 1 visual word (dictionary with 256 words, stop list of size 10); CORNER -- BRISK descriptors, 1 visual word (dictionary with 128 words, stop list of size 10); 3000 TARF descriptors per image; RANSAC verification, 10 iterations; 100K distracting images in the search index.

\begin{enumerate}

\item Number of distracting images

Fig.~\ref{fig_AP_vs_indexSize} shows the dependence of AP on the number of distracting images in the search index.
It is a key feature of TARF that the number of false positive candidates is very low, so even without geometric verification by RANSAC the precision remains high for a large number of distracting images.

\begin{figure}
\begin{tikzpicture}
\begin{axis}[
	xlabel={index size},
	ylabel={AP},
    symbolic x coords={2K, 10K, 100K},
    xtick=data,
    legend style={at={(1.1, 1.0)}, anchor=north west},
]
\addplot coordinates {
	(2K, 0.998)
	(10K, 0.997)
	(100K, 0.995)
};
\addplot coordinates {
	(2K, 0.998)
	(10K, 0.997)
	(100K, 0.996)
};
\addplot coordinates {
	(2K, 0.959)
	(10K, 0.945)
	(100K, 0.908)
};
\addplot coordinates {
	(2K, 0.981)
	(10K, 0.976)
	(100K, 0.961)
};

\addplot coordinates {
	(2K, 0.995)
	(10K, 0.995)
	(100K, 0.995)
};
\addplot coordinates {
	(2K, 0.997)
	(10K, 0.997)
	(100K, 0.997)
};
\addplot coordinates {
	(2K, 0.935)
	(10K, 0.934)
	(100K, 0.933)
};
\addplot coordinates {
	(2K, 0.963)
	(10K, 0.963)
	(100K, 0.963)
};

\legend{
  downscale 30k,
  rotation $30\deg$,
  crop $70\%$,
  blur with $\sigma=4.0$,
  downscale 30k RANSAC,
  rotation $30\deg$ RANSAC,
  crop $70\%$ RANSAC,
  blur with $\sigma=4.0$ RANSAC
}
\end{axis}
\end{tikzpicture}
\caption{
Average precision as a function of the size of the index (i.e. number of distracting images).
}
\label{fig_AP_vs_indexSize}
\end{figure}
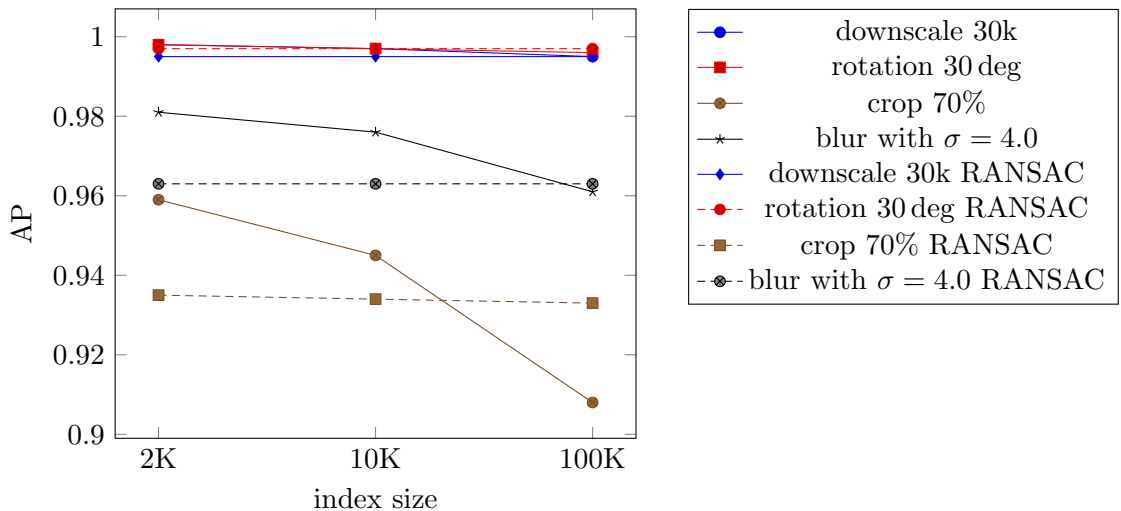

\item Number of dictionaries of visual words

Fig.~\ref{fig_AP_vs_numberOfVW} shows the dependence of AP on the number of dictionaries of visual words.
2x2x2 -- two dictionaries for BLOB descriptors, and two dictionaries for CORNER descriptors, 8 tables in the inverted index;
1x2x2 -- one dictionary for BLOB descriptors, and two dictionaries for CORNER descriptors, 4 tables in the inverted index;
2x1x1 -- 2 tables in the inverted index; 1x1x1 -- 1 table in the inverted index.

Best results are achieved for the 2x2x2 configuration, and the 1x2x2 configuration is nearly as good, thus one visual word is sufficient for BLOB point representation. However, using two visual words to represent each CORNER point is useful and increases AP.

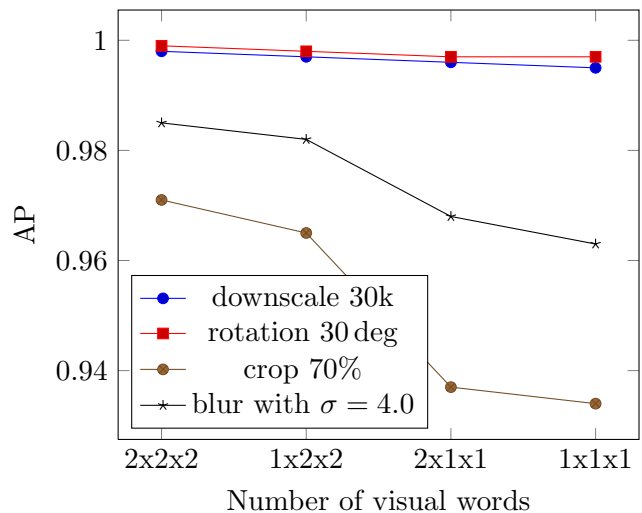
\begin{figure}
\begin{tikzpicture}
\begin{axis}[
	xlabel={Number of visual words},
	ylabel={AP},
    symbolic x coords={2x2x2, 1x2x2, 2x1x1, 1x1x1},
    xtick=data,
    legend style={at={(0.025, 0.025)}, anchor=south west},
]
\addplot coordinates {
	(2x2x2, 0.998)
	(1x2x2, 0.997)
	(2x1x1, 0.996)
	(1x1x1, 0.995)
};
\addplot coordinates {
	(2x2x2, 0.999)
	(1x2x2, 0.998)
	(2x1x1, 0.997)
	(1x1x1, 0.997)
};
\addplot coordinates {
	(2x2x2, 0.971)
	(1x2x2, 0.965)
	(2x1x1, 0.937)
	(1x1x1, 0.934)
};
\addplot coordinates {
	(2x2x2, 0.985)
	(1x2x2, 0.982)
	(2x1x1, 0.968)
	(1x1x1, 0.963)
};
\legend{downscale 30k, rotation $30\deg$, crop $70\%$, blur with $\sigma=4.0$}
\end{axis}
\end{tikzpicture}
\caption{
Average precision as a function of the number of visual words.
}
\label{fig_AP_vs_numberOfVW}
\end{figure}

\item Types of descriptors used

Fig.~\ref{fig_AP_vs_detectorTypes} illustrates the dependence of AP on the type of detectors/descriptors used to build TARF. For BLOB points we have tested SIFT, SURF, SURF detector + BRISK descriptor, and AKAZE~\cite{Alcantarilla2013akaze}. For CORNER points we have always used BRISK. With the only exception of AKAZE feature points, which appear to give low recall for crop 70\%, all other configurations give similar results.

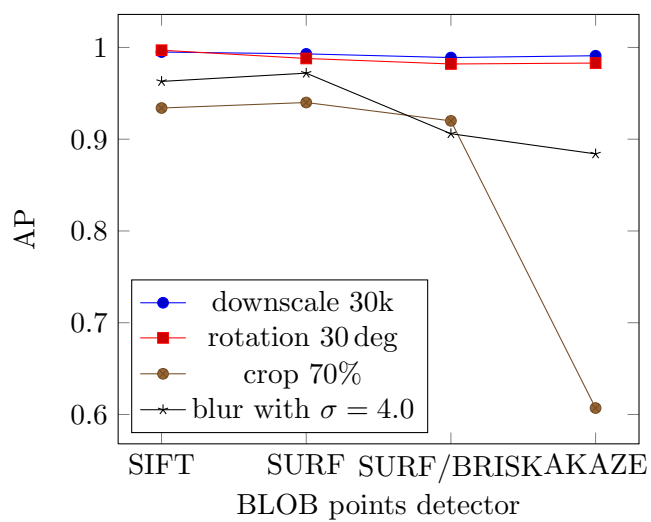
\begin{figure}
\begin{tikzpicture}
\begin{axis}[
	xlabel={BLOB points detector},
	ylabel={AP},
    symbolic x coords={SIFT, SURF, SURF/BRISK, AKAZE},
    xtick=data,
    legend style={at={(0.025, 0.025)}, anchor=south west},
]
\addplot coordinates {
	(SIFT, 0.995)
	(SURF, 0.993)
    (SURF/BRISK, 0.989)
	(AKAZE, 0.991)
};
\addplot coordinates {
	(SIFT, 0.997)
	(SURF, 0.988)
	(SURF/BRISK, 0.982)
	(AKAZE, 0.983)
};
\addplot coordinates {
	(SIFT, 0.934)
	(SURF, 0.94)
	(SURF/BRISK, 0.92)
	(AKAZE, 0.607)
};
\addplot coordinates {
	(SIFT, 0.963)
	(SURF, 0.972)
	(SURF/BRISK, 0.906)
	(AKAZE, 0.884)
};
\legend{downscale 30k, rotation $30\deg$, crop $70\%$, blur with $\sigma=4.0$}
\end{axis}
\end{tikzpicture}
\caption{
Average precision vs types of detectors and descriptors of BLOB feature points.
}
\label{fig_AP_vs_detectorTypes}
\end{figure}

\item Number of TARF descriptors per image

Fig.~\ref{fig_AP_vs_numberOfTarfs} shows the dependence of AP on the number of TARF descriptors extracted from each image; the number of descriptors varies from 1000 to 4000. This graph justifies the choice of 3000 as a default number of TARF descriptors extracted from an image.

\begin{figure}
\begin{tikzpicture}
\begin{axis}[
	xlabel={Number of TARFs per image},
	ylabel={AP},
    xtick=data,
    legend style={at={(0.975, 0.025)}, anchor=south east},
]
\addplot coordinates {
	(1000, 0.992)
	(2000, 0.995)
	(3000, 0.995)
	(4000, 0.995)
};
\addplot coordinates {
	(1000, 0.996)
	(2000, 0.997)
	(3000, 0.997)
	(4000, 0.997)
};
\addplot coordinates {
	(1000, 0.778)
	(2000, 0.909)
	(3000, 0.934)
	(4000, 0.939)
};
\addplot coordinates {
	(1000, 0.947)
	(2000, 0.96)
	(3000, 0.963)
	(4000, 0.964)
};
\legend{downscale 30k, rotation $30\deg$, crop $70\%$, blur with $\sigma=4.0$}
\end{axis}
\end{tikzpicture}
\caption{
Average precision as a function of the number of TARFs extracted from each image.
}
\label{fig_AP_vs_numberOfTarfs}
\end{figure}
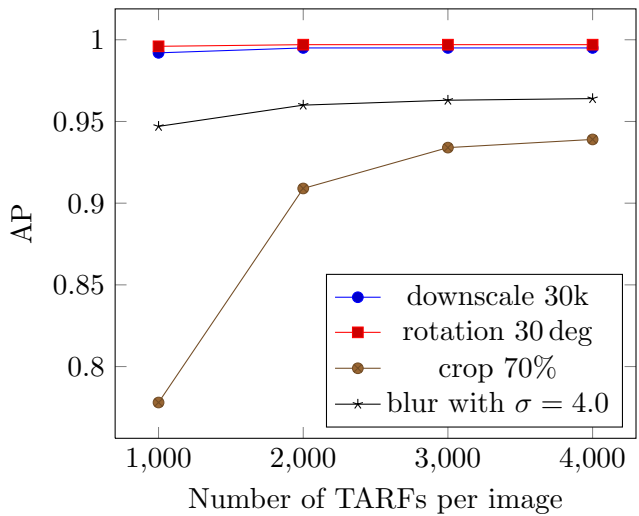

\end{enumerate}

Table~\ref{table_holidays_copydays} shows experimental results for AP on popular \textbf{Holidays} and \textbf{Copydays} datasets~\cite{Jegou2008hammingEmbedding}, with 100K distracting images.
The AP for \textbf{Holidays} and \textbf{Copydays strong} is low due of the low recall. Low recall on these datasets can be explained by the fact that TARF descriptors are matched only if the part of the image is reproduced almost exactly, up to scale/rotation. So, TARF descriptors are very sensitive to the viewpoint change, geometrical distortions of an image, etc. \textbf{Holidays} and \textbf{Copydays strong} image sets include many duplicates which don't have sufficiently large undistorted areas.

\begin{table}
\begin{tabular}{|c|c|}
\hline
image set & average precision (AP) \\
\hline
Holidays & 0.278 \\
Copydays, strong & 0.466 \\
Copydays, crop 30 & 1.0 \\
Copydays, crop 70 & 0.916 \\
Copydays, crop 80 & 0.697 \\
Copydays, jpeg 50 & 1.0 \\
Copydays, jpeg 30 & 0.999 \\
Copydays, jpeg 10 & 0.98 \\
\hline
\end{tabular}
\caption{Average precision for commonly used datasets}
\label{table_holidays_copydays}
\end{table}

\section{Conclusion}

In this paper we have proposed a novel composite image feature -- TARF, which is a combination of three local features. This feature is highly discriminative, because 1) TARF incorporates three local descriptors, thus the probability of a false positive match of a TARF is of the order of a third power of the probability of a false positive match for a single descriptor, which is a very low probability. In other words, it is very improbable that a random combination of three descriptors would match a given TARF. 2) TARF embeds the geometric relationships between three constituent local features.

A method to effectively build an inverted index of TARF features is proposed. Each combination of three descriptors is represented by one 32-bit integer, and geometric information is added to the search index as auxiliary data.

It is shown that this approach makes it possible to build a system for large scale near-duplicates search, with high recall and a low false positives rate.

\bibliographystyle{unsrt}
\bibliography{references}

\end{document}